\documentclass{svproc}
\usepackage{amsmath}
\usepackage{amssymb}
\usepackage{mathtools}
\usepackage{multirow}
\usepackage{booktabs}
\usepackage[T1]{fontenc}

\usepackage{url}

\newcommand{\eg}{e.\,g., }
\newcommand{\ie}{i.\,e., }

\begin{document}
\mainmatter              % start of a contribution
\title{Image Keypoint Matching using Graph Neural Networks}
\titlerunning{Image Keypoint Matching using Graph Neural Networks}  % abbreviated title (for running head)
%                                     also used for the TOC unless
%                                     \toctitle is used
%
\author{Nancy Xu\inst{1} \and Giannis Nikolentzos \inst{2} \and Michalis Vazirgiannis \inst{1,3} \and Henrik Boström \inst{1}}
\authorrunning{Xu et al.} % abbreviated author list (for running head)
%
%%%% list of authors for the TOC (use if author list has to be modified)
\tocauthor{Nancy Xu, Giannis Nikolentzos, Michalis Vazirgiannis, and Henrik Boström}
\institute{KTH Royal Institute of Technology, Stockholm, Sweden\\
\email{\{nancyx,mvaz,bostromh\}@kth.se}
\and
Athens University of Economics and Business, Athens, Greece\\
\email{nikolentzos@aueb.gr}
\and
\'Ecole Polytechnique, Palaiseau, France}

\maketitle              % typeset the title of the contribution

\begin{abstract}
Image matching is a key component of many tasks in computer vision and its main objective is to find correspondences between features extracted from different natural images. When images are represented as graphs, image matching boils down to the problem of graph matching which has been studied intensively in the past. In recent years, graph neural networks have shown great potential in the graph matching task, and have also been applied to image matching. In this paper, we propose a graph neural network for the problem of image matching. The proposed method first generates initial soft correspondences between keypoints using localized node embeddings and then iteratively refines the initial correspondences using a series of graph neural network layers. We evaluate our method on natural image datasets with keypoint annotations and show that, in comparison to a state-of-the-art model, our method speeds up inference times without sacrificing prediction accuracy.
\keywords{keypoint matching, graph neural networks, graph matching}
\end{abstract}

\section{Introduction}
Image matching, the task of finding correspondences between the key features extracted from one image and those extracted from another image of the same object, is a fundamental task in computer vision. Indeed, image matching lies at the heart of several applications in computer vision including 3D structure recovery, simultaneous localization and mapping (SLAM), and change detection, just to name a few. Over the past decades, a large amount of research was devoted to image matching and led to the development of a diverse set of approaches \cite{ma2021image}. 

The pipeline that is typically followed by most image matching algorithms consists of the extraction of a set of key features (or keypoints) from each image and the generation of descriptors for those features. The above steps are followed by some matching approach which is responsible for establishing correspondences between the keypoints of the two images. Recently, the last part of the pipeline, \ie the task of keypoint matching in natural images, has been formulated as a graph matching problem and has been addressed using graph neural network architectures \cite{zanfir2018deep,superglue,consensus}. Images are represented as graphs where nodes correspond to keypoints and edges capture proximity or other types of relations between keypoints. In fact, the problem of graph matching has been studied a lot in the past, but mainly from an algorithmic viewpoint \cite{conte2004thirty}. Unfortunately, most derived formulations correspond to NP-hard problems, and therefore, are not useful for most real-world settings. With the advent of deep learning and the success of graph neural networks, the problem of learning how to compute correspondences of nodes (or of other structures) between two graphs attracted increasing attention \cite{zanfir2018deep,heimann2018regal,wang2019learning,bai2019simgnn}. Though not exact, approaches that utilize graph neural networks to find correspondences between nodes are very efficient, while they also learn representations that are best suited for the task at hand.

In this paper, we capitalize on recent advances in graph representation learning, and we propose a graph neural network for establishing meaningful structural correspondences of keypoints between two natural images. More specifically, we take as starting point the Deep Graph Matching Consensus (DGMC) method proposed in \cite{consensus}. DGMC operates in two stages, a local feature matching stage which generates an initial soft correspondence matrix, and a neighborhood consensus stage which iteratively refines the soft correspondence matrix by repeatedly applying a single graph neural network to the scores. The final correspondence matrix is a combination of the initial soft matrix and the final result of the iterations.

We develop a new approach for obtaining and combining soft correspondence matrices that is more intuitive and more attractive in terms of running time. The proposed method first generates initial soft correspondences between keypoints using localized node embeddings, and then iteratively refines the initial correspondences using a series of graph neural network layers. The final correspondence matrix is formed of all the matricies produced during the consensus stage. This change increases the complexity of the neighborhood consensus stage which reduces the number of iterations needed, thus decreasing inference time. Experiments on standard keypoint matching datasets demonstrate that in comparison to the DGMC, the proposed method is in most cases more efficient, while in terms of performance, it matches or outperforms the baseline model.

The rest of this paper is organized as follows.
Section~\ref{sec:related_work} provides an overview of the related work.
Section~\ref{sec:methodology} provides a detailed description of the proposed model.
Section~\ref{sec:experiments} evaluates the proposed model in the task of keypoint matching in natural images.
Finally, section~\ref{sec:conclusion} concludes.

\section{Related Work}\label{sec:related_work}
Traditional methods for identifying correspondences between images typically deal with instance matching tasks where correspondences are found between two images of the same scene or object. These methods use representations such as image pyramids which are formed by repeatedly filtering and subsampling a given image. Image features are computed densely over image pyramids and can be sampled densely \cite{HOG} or used to identify keypoints \cite{SIFT}. Notably, SIFT-Flow combines the image pyramid with optical flow methods for instance matching \cite{siftflow}. 
Unfortunately, most traditional methods depend on handcrafted features which must be manually defined. 

In comparison, deep learning methods automatically learn important image features and are widely used in computer vision. For instance, Long \textit{et al.} applied in \cite{CNN-basic} a straightforward convolutional neural network to image matching and achieved similar and in some cases, superior results to those of SIFT-Flow. The introduction of datasets specialized for image matching such as Proposal Flow (PASCAL-PF) \cite{pascalpf}, WILLOW-ObjectClass \cite{willow}, and the PASCAL-VOC Berkeley annotations \cite{berkeley}, greatly improved the ease of developing new deep learning methods. Examples include FCSS, which uses a convolutional self-similarity layer to encode local self-similarity patterns \cite{FCSS} and SFNet which outputs a semantic flow field using a CNN \cite{SFN}. 

Finding correspondences between keypoints in images can also be seen as a graph matching problem. Graph matching can be formulated as a quadratic assignment problem, which turns out to be a notoriously difficult problem \cite{quadsurvey}. A more computationally tractable formulation is to treat the problem as an optimal transport problem. Xu \textit{et al.} demonstrated in \cite{graphoptimal} the success of using this formulation along with the Sinkhorn algorithm for graph matching \cite{sinkhorn}. 
Graph neural networks can effectively and automatically generate node or graph representations using message passing schemes \cite{gnns}. Several recent methods use a combination of graph neural networks and the Sinkhorn algorithm for graph matching \cite{zanfir2018deep,wang2019learning,superglue,consensus}. By using a graph neural network to generate similarity scores followed by the application of the Sinkhorn normalization, we can build an end-to-end trainable framework for semantic matching between keypoints extracted from images.

\section{Graph Neural Networks for Image Keypoint Matching}\label{sec:methodology}
Before presenting the proposed method, we begin by introducing some key notation for graphs which will be used later. Let $G=(V,E)$ be a graph where $V$ denotes the set of nodes and $E$ the set of edges. We will denote by $n$ the number of vertices and by $m$ the number of edges. Let also $\mathbf{A} \in \mathbb{R}^{n \times n}$ denote the adjacency matrix of $G$. Following previous work \cite{consensus}, we represent images as graphs where nodes correspond to keypoints and edges are obtained via the Delaunay triangulation of the keypoints. Furthermore, every node in the graph is associated with a feature vector (\ie the input features of keypoints), and we use $\mathbf{X} \in \mathbb{R}^{n \times d}$ to denote those features where $d$ is the feature dimensionality. The edges of the graph can also be potentially associated with a feature vector (\eg distance of two endpoints), and we use $\mathbf{E} \in \mathbb{R}^{n \times n \times d}$ to denote those features.

Given a source and a target image, we fist map these images to graphs $G_{s}$ and $G_{t}$, respectively. Let $n_s$ and $n_t$ denote the number of nodes of the two graphs. Without loss of generality, we assume that $n_s \leq n_t$. The goal of the method is to establish correspondences between $G_{s}$ and $G_{t}$ and match two pairs of keypoints. 

Similar to DGMC \cite{consensus}, our method consists of a local feature matching stage followed by a neighborhood consensus stage, which are both presented in the following subsections. The local feature matching stage finds initial correspondences between images based on the similarity of their node embeddings and is implemented as a graph neural network layer $\text{GNN}^{(0)}$. The neighborhood consensus stage then refines these initial correspondences which we implement as a series of graph neural network layers $\text{GNN}^{(1)},\ldots,\text{GNN}^{(K)}$. 

\subsection{Local Feature Matching}
Starting with $G_{s}$ and $G_{t}$, the representations of the nodes are updated by feeding both graphs into a graph neural network, \ie $\mathbf{H} = \text{GNN}^{(0)}(\mathbf{A}, \mathbf{X}, \mathbf{E})$. Since we use a graph neural network to update the features of the keypoints, we obtain localized and permutation equivariant representations. Then, the similarity between keypoints extracted from the two images is determined by the similarity of their node representations $\mathbf{H}_{s}$ and $\mathbf{H}_{t}$. Specifically, we generate matrix $\mathbf{K}^{(0)}$ as follows $\mathbf{K}^{(0)} = \mathbf{H}_{s} \, \mathbf{H}_{t}^\top$, and then, we can obtain initial soft correspondences between the keypoints of the two images via the Sinkhorn algorithm \cite{cuturi2013sinkhorn}: 
\begin{equation*}
    \mathbf{S}^{(0)} = \text{sinkhorn}\big(\mathbf{K}^{(0)}\big) \in [0,1]^{n_s \times n_t}
\end{equation*}
The Sinkhorn normalization produces a rectangular doubly-stochastic correspondence matrix such that $\sum_{j=1}^{n_t} \mathbf{S}_{i,j}^{(0)} = 1, \, \forall i \in \{ 1,\ldots,n_s\}$ and $\sum_{i=1}^{n_s} \mathbf{S}_{i,j}^{(0)} \leq 1, \, \forall j \in \{1,\ldots,n_t\}$.

As discussed in \cite{consensus}, the Sinkhorn algorithm is expensive in terms of running time, while it also might lead to vanishing gradients. Furthermore, it may also converge to inconsistent solutions which the next stage of the algorithm might find difficult to refine. Therefore, instead of applying Sinkhorn, the problem is partially relaxed by dropping the following constraint $\sum_{i=1}^{n_s} \mathbf{S}_{i,j}^{(0)} \leq 1, \, \forall j \in \{1,\ldots,n_t\}$, and the initial correspondence matrix is computed by applying row-wise softmax normalization on matrix $\mathbf{K}^{(0)}$ as follows:
\begin{equation*}
    \mathbf{S}^{(0)} = \text{softmax}\big( \mathbf{K}^{(0)} \big) 
\end{equation*}

\subsection{Neighborhood Consensus} 

The initial correspondences only take local features into account, which makes it more prone to falsely matching nodes which are only superficially similar. The neighborhood consensus step aims to detect these false correspondences and iteratively resolve them. The iteration is done by applying a series of graph neural network layers which we refer to as $\text{GNN}^{(1)},\ldots,\text{GNN}^{(K)}$. In contrast to DGMC proposed in \cite{consensus}, which uses a shared graph neural network for all iterations, in our case no weight sharing is performed and each graph neural network consists of its own trainable parameters. The idea behind this change is to increase the complexity of the neighborhood consensus stage while increasing efficiency by reducing the number of iterations needed to refine the correspondence matrices.

At each iteration, a graph neural network layer produces a new correspondence matrix by updating the previous correspondence matrix. This process is applied $k$ times resulting in $\mathbf{S}^{(1)}, \ldots, \mathbf{S}^{(K)}$.  

For every iteration $k$, where $k \in \{1,\ldots,K\}$, each consensus loop begins by assigning an arbitrary coloring to the nodes of $G_s$ using randomly generated vectors. Let $\mathbf{R}_s$ be a matrix whose $i$-th row contains the vector of the $i$-th node of $G_s$. Then, the model assigns a coloring to the nodes of $G_t$ based on their relationship to the nodes of $G_s$ (as captured by matrix $\mathbf{S}^{(k-1)}$). Thus, matrix $\mathbf{R}_t$ is produced as follows: $\mathbf{R}_t = \big[\mathbf{S}^{(k-1)}\big]^\top \, \mathbf{R}_s$. These colors are distributed across their respective graphs by using synchronous message passing, \ie by applying $\text{GNN}^{(k)}$ to each graph. That is, the model computes new matrices of features $\mathbf{Z}_s^{(k)} = \text{GNN}^{(k)}(\mathbf{A}_s, \mathbf{R}_s, \mathbf{E}_s)$ and $\mathbf{Z}_t^{(k)} =\text{GNN}^{(k)}(\mathbf{A}_t, \mathbf{R}_t, \mathbf{E}_t)$. By comparing the emerging representations of two nodes, a matrix $\mathbf{D} \in \mathbb{R}^{n_s \times n_t}$ is produced which measures the neighborhood consensus between all pairs of nodes. For example, for the $i$-th node of $G_s$ and the $j$-th node of $G_t$, we have that:
\begin{equation*}
    \mathbf{D}_{i,j} =  \big[\mathbf{Z}_s^{(k)}\big]_{i,:} - \big[\mathbf{Z}_t^{(k)}\big]_{j,:}
\end{equation*}
where $\big[\mathbf{Z}\big]_{i,:}$ denotes the $i$-th row of matrix $\mathbf{Z}$. Matrix $\mathbf{D}$ is then transformed using a multi-layer perceptron and the new correspondence matrix is computed as:
\begin{equation*}
    \mathbf{S}^{(k)} = \text{softmax}\big(\mathbf{S}^{(k-1)} + \text{MLP}(\mathbf{D})\big)
\end{equation*}
Instead of considering only two correspondence matrices as in \cite{consensus}, in our setting, the final correspondence matrix $\mathbf{S}$ is computed as the weighted sum of all the soft correspondence matrices $\mathbf{S}^{(0)},\ldots,\mathbf{S}^{(K)}$, where the weights correspond to trainable parameters followed by a final softmax function (applied in a row-wise manner).
\begin{equation*}
    \mathbf{S} = \text{softmax} \Big( \mathbf{w}_0 \, \mathbf{S}^{(0)} + \mathbf{w}_1 \, \mathbf{S}^{(1)} + \ldots + \mathbf{w}_k \, \mathbf{S}^{(K)} \Big)
\end{equation*}
where $\mathbf{w} \in \mathbb{R}^k$ is a trainable vector. To train the model, we employ the categorical cross-entropy:
\begin{equation*}
    \mathcal{L} =  -\sum_{i=1}^{n_s} \log \big( \mathbf{S}_{i,\pi(i)} \big)
\end{equation*}
where $\pi(i)$ gives the node of $G_t$ that matches to node $i$ of $G_s$. Since the loss function is differentiable, we can train the model in an end-to-end fashion using stochastic gradient descent. 

\section{Experiments}\label{sec:experiments}
In this section, we report on the experimental evaluation of the proposed method and compare with state-of-the-art methods.

\subsection{Datasets}
We evaluate the proposed method in the task of keypoint matching in natural images and in the task of geometric feature matching, where only the coordinates of the keypoints are available and not their visual features. For the first task, we experiment with the following two datasets: (i) PASCAL-VOC \cite{pascal} with Berkeley annotations of keypoints \cite{berkeley}; and (ii) WILLOW-ObjectClass \cite{willow}. The PASCAL-VOC dataset is an extension of PASCAL-VOC 2011 and contains annotations of body parts for 20 semantic classes. The Willow ObjectClass dataset consists of 5 categories from Caltech-256 and Pascal VOC 2007. For the second task, we experiment with the PASCAL-PF dataset \cite{pascalpf}. The PASCAL-PF dataset consists of annotations for 20 semantic classes. The annotations for each image pair consist of a set of keypoint coordinate pairs where each pair shares the same arbitrary numeric label.  

\subsection{Baselines}
We compare the proposed model against recently proposed neural network models that have achieved stat-of-the-art performance. More specifically, on the PASCAL-VOC and WILLOW ObjectClass datasets, we compare the proposed model against the following three models: (i) GMN \cite{zanfir2018deep}; (ii) PCA-GM \cite{wang2019learning}; and (iii) DGMC \cite{consensus}. On the PASCAL-PF dataset, we compare our model against two models: (i) GMN \cite{zhang2019deep}; and (ii) DGMC \cite{consensus}.
For the DGMC model, $L$ refers to the number of neighborhood consensus iterations used. For all baselines, we use the results reported in \cite{consensus}.

\subsection{Experimental Setup} 
On all three considered datasets, we follow the experimental setups of previous studies \cite{zanfir2018deep,wang2019learning,consensus,choy2016universal}, and we use the same training and test splits as in those studies.

For the proposed model, we set the number of epochs to $30$, and the batch size to $512$. We use the Adam optimizer with a learning rate of $10^{-3}$. We set the hidden-dimension size of the graph neural networks to $128$. We tune the following two hyperparameters: (i) the number of graph neural networks used in the neighborhood consensus step $K$; and (ii) the number of message passing iterations of each graph neural network $r$. For the proposed model and DGMC, two types of edge features are evaluated: (i) isotropic edge features that use normalized relative distances which result in the neighbors of a node being equally considered during message passing; and (ii) anisotropic edge features that use 2D cartesian coordinates which result in some neighbors having a greater weight during message passing.
$\text{GNN}^{(0)}, \ldots,\text{GNN}^{(K)}$ all correspond to SplineCNN models \cite{spline}. The message passing scheme employed by the SplineCNN model is defined as:
\begin{equation*}
    \mathbf{h}_i^{(t+1)} = \sigma \Big( \mathbf{W}^{(t+1)} \mathbf{h}_i^{(t)} + \sum_{j \in \mathcal{N}(i)} \mathbf{\Phi}_\theta^{(t+1)} (\mathbf{e}_{i,j}) \mathbf{h}_j^{(t)}  \Big)
\end{equation*}
where $\mathcal{N}(i)$ denotes the set of neighbors of the $i$-th node of the graph, and the trainable B-spline based kernel function $\mathbf{\Phi}_\theta(\cdot)$ is conditioned on edge features.

The proposed model is implemented in PyTorch \cite{pytorch} using the PyTorch Geometric library \cite{pytorch-geo} and KeOPs \cite{keops}. All experiments were performed on a single machine equipped with NVidia Tesla P100-PCIE GPU. 

\begin{table*}[t]
  \centering
  \caption{Hits@1 (\%) on the PASCAL-PF and PASCAL-VOC dataset}
  \label{tab:pascalpf}
  \def\arraystretch{1.2}
  \resizebox{\textwidth}{!}{
  \begin{tabular}{ccccccccccccccccccccc|c}
    \toprule
    PASCAL-VOC & Aero & Bike & Bird & Boat & Bottle & Bus & Car & Cat & Chair & Cow & Table & Dog & Horse & M-Bike & Person & Plant & Sheep & Sofa & Train & TV & Mean\\
    \midrule
    GMN & 31.1 & 46.2 & 58.2 & 45.9 & 70.6 & 76.5 & 61.2 & 61.7 & 35.5 & 53.7 & 58.9 & 57.5 & 56.9 & 49.3 & 34.1 & 77.5 & 57.1 & 53.6 & 83.2 & 88.6 & 57.9 \\
    PCA-GM & 40.9 & 55.0 & 65.8 & 47.9 & 76.9 & 77.9 & 63.5 & 67.4 & 33.7 & 66.5 & 63.6 & 61.3 & 58.9 & 62.8 & 44.9 & 77.5 & 67.4 & 57.5 & 86.7 & 90.9 & 63.8\\
    
    \hline 
    DGMC $L = 20$ Iso & 50.1 & 65.4 & 55.7 & 65.3 & 80.0 & 83.5 & 78.3 & 69.7 & 34.7 & 60.7 & 70.4 & 59.9 & 70.0 & 62.2 & 56.1 & 80.2 & 70.3 & 88.8 & 81.1 & 84.3 & 68.3\\
    Ours $K = 1$, $r = 4$ Iso & 47.8 & 66.5 & 57.5& 56.9 & 74.4 & 85.1 & 75.7 & 69.4 & 33.8 & 55.9 & 80.9 & 63.1 & 69.5 & 58.0 & 49.8 & 78.6 & 61.5 & 90.2 & 77.6 & 83.1 & 66.8\\
    
    \hline
    DGMC $L = 20$ Aniso & 47.0 & 65.7 & 56.8 & 67.6 & 86.9 & 87.7 & 85.3 & 72.6 & 42.9 & 69.1 & 84.5 & 63.8 & 78.1 & 55.6 & 58.4 & 98.0 & 68.4 & 92.2 & 94.5 & 85.5 & 73.0\\
    Ours $K = 1$, $r = 5$ Aniso & 52.0 & 71.5 & 56.0 & 72.1 & 84.6 & 85.5 & 83.9 & 75.7 & 48.0 & 70.5 & 88.1 & 66.0 & 85.1 & 64.5 & 58.2 & 96.2 & 67.5 & 93.5 & 93.6 & 88.7 & 75.0\\
  \bottomrule
  \end{tabular}}\\
  \vspace{.2cm}
  \resizebox{\textwidth}{!}{
  \begin{tabular}{ccccccccccccccccccccc|c}
    \toprule
    PASCAL-PF & Aero & Bike & Bird & Boat & Bottle & Bus & Car & Cat & Chair & Cow & Table & Dog & Horse & M-Bike & Person & Plant & Sheep & Sofa & Train & TV & Mean\\
    \midrule
    (Zhang \& Lee, 2019) & 76.1 & 89.8 & 93.4 & 96.4 & 96.2 & 97.1 & 94.6 & 82.8 & 89.3 & 96.7 & 89.7 & 79.5 & 82.6 & 83.5 & 72.8 & 76.7 & 77.1 & 97.3 & 98.2 & 99.5 & 88.5 \\
    DGMC L = 20 & 81.1 & 92.0 & 94.7 & 100.0 & 99.3 & 99.3 & 98.9 & 97.3 & 99.4 & 93.4 & 100.0 & 99.1 & 86.3 & 86.2 & 87.7 & 100.0 & 100.0 & 100.0 & 100.0 & 99.3 & 95.7\\
    Ours $K$ = 5, $r = 4$ & 82.2 & 92.2 & 94.5 & 99.4 & 99.3 & 98.4 & 98.6 & 98.2 & 98.6 & 94.9 & 100.0 & 99.0 & 85.6 & 87.2 & 87.8 & 99.5 & 100.0 & 99.4 & 99.5 & 98.9 & 95.7\\ 
    \toprule
  \end{tabular}}\\
\end{table*}

\subsection{Results} 
Following previous work, we use Hits@$1$ as our evaluation metric, which measures the proportion of correct matches in the top result. We also measure the running time of the proposed model.

\paragraph{PASCAL-VOC.} 
Table~\ref{tab:pascalpf} (top) shows the Hits@1 scores achieved by the different methods. We observe that the anisotropic variants of the proposed model and of DGMC outperform all the other models by wide margins. Interestingly, the anisotropic variant of the proposed model is the best performing method since it yields a relative increase of $2\%$ on the mean Hits@1 over DGMC. In terms of running time, as shown in Table~\ref{tab:pf-inf} (top), the variants of the proposed model are faster than those of DGMC both during training and during inference. This can be of high importance for real-time scenarios where generating predictions in a small amount of time is of a critical nature.

\begin{table}[t]
  \centering
  \caption{Average inference and training times on the Pascal-PF and PASCAL-VOC datasets.}
  \label{tab:pf-inf}
  \scriptsize
  \def\arraystretch{1.2}
  \begin{tabular}{l|c|c}
    \toprule
    PASCAL-VOC & Inference Time & Training Time (per epoch)\\
    \midrule
    DGMC $L = 20$ Iso & 89ms & 29s\\
    Ours $K = 1$, $r = 4$ Iso & 36ms & 17s\\
    \hline
    DGMC $L = 20$ Aniso & 158ms & 60s\\
    Ours $K = 1$, $r = 5$ Aniso & 94ms & 47s\\
  \bottomrule
  \end{tabular}\\
  \vspace{.2cm}
  \begin{tabular}{l|c|c}
    \toprule
    PASCAL-PF & Inference Time & Training Time (per epoch)\\
    \midrule
    DGMC $L = 20$ & 40ms & 34s\\
    Ours $K=5$, $r=4 \qquad \, \, \,$ & 19ms & 36s\\
  \bottomrule
  \end{tabular}
\end{table}

% \begin{table}[t]
%   \centering
%   \caption{Average inference and training times on the Pascal-PF and PASCAL-VOC datasets.}
%   \label{tab:pf-inf}
%   \scriptsize
%   \def\arraystretch{1.2}
%   \begin{tabular}{l|c}
%     \toprule
%     PASCAL-VOC & Inference Time\\
%     \midrule
%     DGMC $L = 20$ Iso & 89ms\\
%     Ours $K = 1$, $r = 4$ Iso & 36ms\\
%     \hline
%     DGMC $L = 20$ Aniso & 158ms\\
%     Ours $K = 1$, $r = 5$ Aniso & 94ms\\
%   \bottomrule
%   \end{tabular}\\
%   \vspace{.2cm}
%   \begin{tabular}{l|c}
%     \toprule
%     PASCAL-PF & Inference Time\\
%     \midrule
%     DGMC $L = 20$ & 40ms\\
%     Ours $K=5$, $r=4 \qquad \, \, \,$ & 19ms\\
%   \bottomrule
%   \end{tabular}
% \end{table}

\begin{figure}[t]
  \centering
  \includegraphics[width=\linewidth]{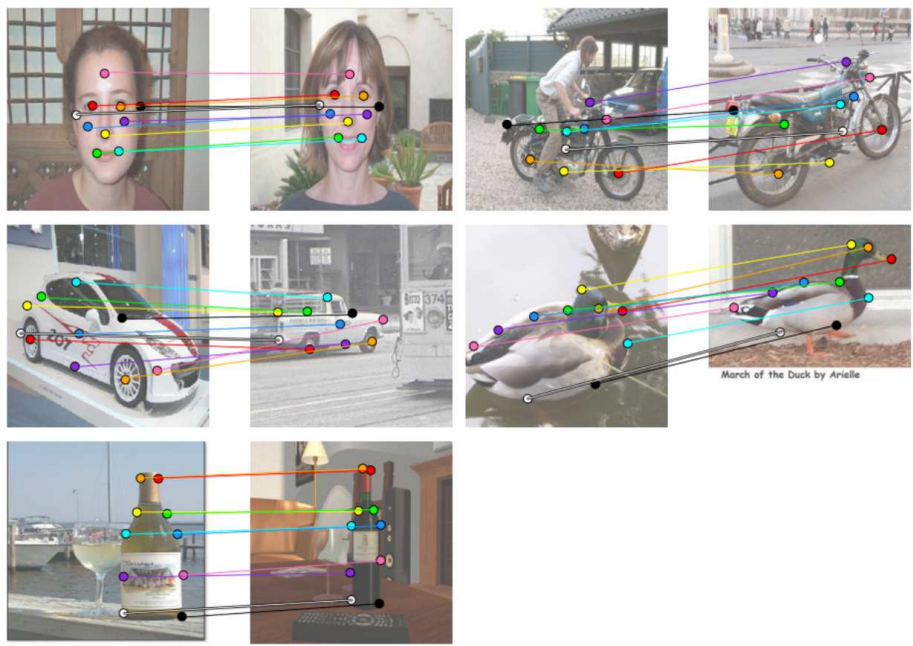}
  \caption{Examples of matching keypoints from the WILLOW-ObjectClass dataset.}
  \label{fig:examples}
\end{figure}

\paragraph{WILLOW-ObjectClass.}
Tables~\ref{tab:will} and~\ref{tab:will-inf} illustrate the performance (Hits@1 score) and running time of the different models, respectively. The anisotropic variant of DGMC outperforms all the other methods, while the anisotropic variant of the proposed model is the second best method and is slightly outperformed by DGMC. Surprisingly, the isotropic instance of our model achieves much smaller Hits@1 score than that of DGMC, while it is even outperformed by PCA-GM. In both cases, the proposed model is trained in a smaller amount of time than DGMC, while it is also more efficient at inference time. In Figure~\ref{fig:examples}, we also provide some examples where the proposed model has successfully find correspondences between keypoints extracted from pairs of images. 

\begin{table*}[t]
  \centering
  \caption{Hits@1 (\%) on the WILLOW-ObjectClass dataset}
  \label{tab:will}
  \def\arraystretch{1.2}
  \scriptsize
  \begin{tabular}{lccccc}
    \toprule
    WILLOW & face & motorbike & car & duck & winebottle\\
    \midrule
    GMN & 99.3  & 71.4 & 74.3 & 82.8 & 76.7 \\
    PCA-GM & 100.0 & 76.7 & 84.0 & 93.5 & 96.9\\
    \hline 
    DGMC $L = 20$ Iso & 100.00 ± 0.00 & 92.05 ± 3.24 & 90.28 ± 4.67 & 88.97 ± 3.49 & 97.14 ± 1.83\\ 
    Ours $K = 3$, $r = 3$ Iso & 99.40 ± 0.46 & 87.78 ± 3.76 & 83.82 ± 4.09 & 85.20 ± 2.86 & 92.53 ± 1.69 \\
    \hline
    DGMC $L = 10$ Aniso & 100.00 ± 0.00 & 98.80 ± 1.58 & 96.53 ± 1.55 & 93.22 ± 3.77 & 99.87 ± 0.31\\
    Ours $K = 3$, $r = 4$ Aniso & 99.85 ± 0.24 & 99.35 ± 0.71 & 97.88 ± 1.09 & 93.17 ± 2.67 & 99.15 ± 0.73\\
  \bottomrule
\end{tabular}
\end{table*}

\begin{table}[t]
  \centering
  \caption{Average inference and training times on the WILLOW-ObjectClass dataset.}
  \label{tab:will-inf}
  \footnotesize
  \def\arraystretch{1.2}
  \begin{tabular}{l|c|c}
    \toprule
    
    WILLOW & Inference Time & Training Time (per Epoch)\\
    \midrule
    DGMC L = 20 Iso & 57ms & 13s\\
    Ours $K = 3$, $r = 3$ Iso & 37ms & 8s \\
    \hline
    DGMC L = 10 Aniso & 64ms & 23s\\
    Ours $K = 3$, $r = 4$  Aniso & 51ms & 16s\\
  \bottomrule
  \end{tabular}
\end{table}

% \begin{table}[t]
%   \centering
%   \caption{Average inference and training times on the WILLOW-ObjectClass dataset.}
%   \label{tab:will-inf}
%   \footnotesize
%   \def\arraystretch{1.2}
%   \begin{tabular}{l|c}
%     \toprule
%     WILLOW & Inference Time\\
%     \midrule
%     DGMC L = 20 Iso & 57ms\\
%     Ours $K = 3$, $r = 3$ Iso & 37ms\\
%     \hline
%     DGMC L = 10 Aniso & 64ms\\
%     Ours $K = 3$, $r = 4$  Aniso & 51ms\\
%   \bottomrule
%   \end{tabular}
% \end{table}

\paragraph{PASCAL-PF.}
Table~\ref{tab:pascalpf} (bottom) reports the Hits@1 score achieved by the proposed model and the baselines. We observe that the proposed model matches the performance of DGMC, while both models outperform the model proposed in \cite{zhang2019deep}. This indicates that our model is very effective even if visual features are not available. Table~\ref{tab:pf-inf} (bottom) illustrates the running time of the proposed model and that of DGMC. The proposed model's average inference time per image is much smaller than that of DGMC, while training times of the two models are similar to each other.

\section{Conclusion}\label{sec:conclusion}
In this paper, we presented a graph neural network for the problem of image matching. Our starting point is a previous work on graph neural networks for graph matching \cite{consensus}, which we modify to use not only two correspondence matrices, but all the matrices that are produced during the consensus stage. Furthermore, we also replace the iteration in the neighborhood consensus step with a series of graph neural networks. Experiments on standard image matching datasets showed that the proposed model is more efficient than the baseline model while it maintains and in some cases improves prediction performance. 

\section{Acknowledgements} 
This research is co-financed by Greece and the European Union (European Social Fund- ESF) through the Operational Programme \guillemotleft Human Resources Development, Education and Lifelong Learning\guillemotright in the context of the project ``Reinforcement of Postdoctoral Researchers - 2\textsuperscript{nd} Cycle'' (MIS-5033021), implemented by the State Scholarships Foundation (IKY). This work was also supported by the Wallenberg AI, Autonomous Systems and Software Program (WASP).

%WASP 7156

% ---- Bibliography ----
\bibliographystyle{splncs03}
\bibliography{biblio.bib}

\end{document}